\documentclass[10pt,twocolumn,letterpaper]{article}

\usepackage{cvpr}

\usepackage{epsfig}
\usepackage{graphicx}

\usepackage{times}
\usepackage{amsmath}
\usepackage{amssymb}

\usepackage{multicol}
\usepackage{multirow}

\usepackage{multicol}
\usepackage[font=small,labelfont=bf,tableposition=bottom,figureposition=bottom]{caption}

\usepackage{authblk}

\newcommand*{\helvetica}{\fontfamily{phv}\selectfont\scriptsize}

\usepackage[breaklinks=true,bookmarks=false]{hyperref}

\cvprfinalcopy 

\def\cvprPaperID{****} 

\begin{document}

\begin{multicols}{1}
\title{How Much Does Audio Matter to Recognize Egocentric Object Interactions?}

\author[1]{Alejandro Cartas}
\author[2]{Jordi Luque}
\author[1]{Petia Radeva}
\author[2]{Carlos Segura}
\author[3]{Mariella Dimiccoli}
\affil[1]{University of Barcelona}
\affil[2]{Telefonica I+D, Research, Spain}
\affil[3]{Institut de Rob\`otica i Inform\`atica Industrial (CSIC-UPC)}

\renewcommand\Authands{ and }

\maketitle

\end{multicols}
\thispagestyle{empty}

\begin{abstract}
Sounds are an important source of information on our daily interactions with objects. For instance, a significant amount of people can discern the temperature of water that it is being poured just by using the sense of hearing\footnote{For more information on this fun fact, please refer to experiment 1 in \cite{velasco2013soundTemperature}}. However, only a few works have explored the use of audio for the classification of object interactions in conjunction with vision or as single modality. In this preliminary work, we propose an audio model for egocentric action recognition and explore its usefulness on the parts of the problem (noun, verb, and action classification). Our model achieves a competitive result in terms of verb classification (34.26\% accuracy) on a standard benchmark with respect to vision-based state of the art systems, using a comparatively lighter architecture.
\end{abstract}

\section{Introduction}

Human experience of the world is inherently multimodal \cite{frassinetti2002enhancement,campbell2007processing}. We employ different senses to perceive new information both passively and actively when we explore our environment and interact with it. In particular, object manipulations almost always have an associate sound (e.g. open tap) and we naturally learn and exploit these associations to recognize object interactions by relying on available sensory information (audio, vision or both). Recognizing object interactions from visual information has a long story of research in Computer Vision \cite{zhang2019survey}. Instead, audio has comparatively been little explored in this context and most works have been focused on auditory scene classification \cite{Li2013,Piczak2015a,Rakotomamonjy2015,Roma2013RECURRENCEQA}. Only more recently, audio has been used in conjunction with visual information for scene classification \cite{aytar2016soundnet} and object interactions recognition \cite{Arandjelovic17,Owens2016VisuallyIS}.

All works previous to the introduction of Convolutional Neural Networks (CNNs) for audio classification \cite{Li2013,Piczak2015a,Rakotomamonjy2015,Roma2013RECURRENCEQA} shared a common pipeline consisting in first extracting the time-frequency representation from the audio signal, such as the mel-spectrogram \cite{Li2013}, and then classifying it with methods like Random Forests \cite{Piczak2015a} or Support Vector Machines (SVMs) \cite{Roma2013RECURRENCEQA}. More recently Rakotomamonjy and Gasso \cite{Rakotomamonjy2015} proposed to work directly on the image of the spectrogram instead of its coefficients to extract audio features. More specifically, they used histogram of gradients (HOGs) of the spectrogram image as features, on the top of which they applied a SVM classifier. The idea of using the image spectrogram as input to a CNN to learn features in a end-to-end fashion was firstly proposed in \cite{Piczak2015b}.

\begin{figure*}[!t]
\begin{center}
\includegraphics[width=0.75\textwidth]{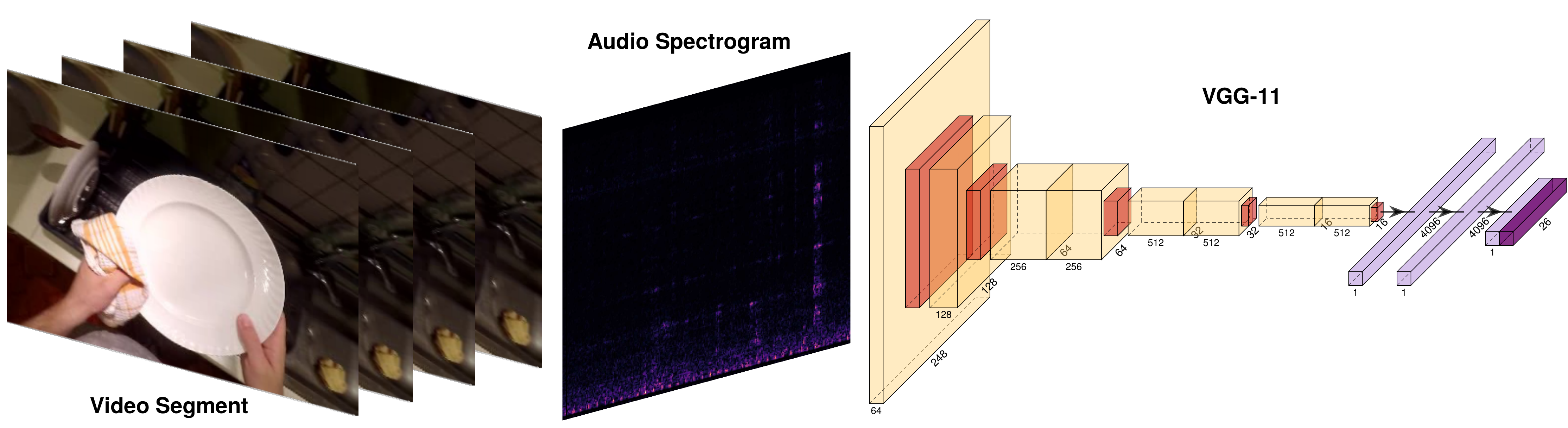}
\caption[]{Audio-based classification overview.}
\label{fig:overview}
\end{center}
\end{figure*}

\begin{figure*}[!t]
\begin{center}
\begin{minipage}{0.15\textwidth}%
\helvetica%
\begin{center}
Video segments time duration stats (secs)\\
\begin{tabular}{r l}
\hline
\textbf{Min}: &\hspace{-2.5mm} 0.5\\
\textbf{Mean}: &\hspace{-2.5mm} 3.38\\
\textbf{Median}: &\hspace{-2.5mm} 1.78\\
\textbf{Std. Dev.}: &\hspace{-2.5mm} 5.04\\
\textbf{Mode}: &\hspace{-2.5mm} 1.0\\
\textbf{Max}: &\hspace{-2.5mm} 145.16\\
\end{tabular}
\end{center}%
\end{minipage}%
\hspace{1em plus 1fill}
\begin{minipage}{0.8\textwidth}
\includegraphics[width=\textwidth]{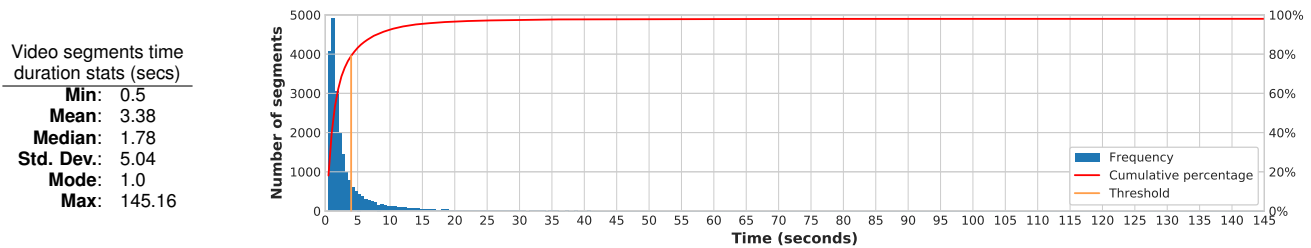}
\end{minipage}
\caption[]{Statistics and histogram of the time duration of the video segments in the EPIC Kitchens dataset training split with a bin size of half a second.}
\label{fig:timeDurationStats}
\end{center}
\end{figure*}

Later on, Owens et al.\cite{Owens2016VisuallyIS} presented a model that takes as input a video without audio capturing a person scratching or touching different materials, and generates the emitting synchronized sounds. In their model, the sequential visual information is processed by a CNN (AlexNet \cite{Krizhevsky2012AlexNet}) connected to a Long Short-Term Memory (LSTM) \cite{Hochreiter1997LSTM}, and the problem is treated as a regression of an audio cochleagram. Aytar et al. \cite{aytar2016soundnet} proposed to perform auditory scene classification by using transfer learning from visual CNNs in an unsupervised fashion. In \cite{Harwath2016UnsupervisedLO}, a two-stream neural network learns semantically meaningful words and phrases at the spectral feature level from a natural input image and its spoken captions without relying on speech recognition or text transcriptions.

Given the indisputable importance of sounds in human and machine perception, this work aims at understanding the role of audio in egocentric action recognition, and in particular to unveil when audio and visual features provide complementary information in the specific domain of egocentric object interaction in a kitchen. 

\section{Audio-based Classification}

Our model is a VGG-11\cite{Simonyan14c} neural network that takes as input an audio spectrogram. This spectrogram only considers the first four seconds of a short video segment. To determine such time interval, we calculated the time duration statistics of the video segments in the filtered training split. We also calculated its time duration histogram and cumulative percentage as shown in colors black and red, respectively in Fig. \ref{fig:timeDurationStats}. As it can be observed from the threshold line in yellow, setting the audio time size window equal to 4 seconds, allows to completely cover $80.697\%$ of all video segments in one window. When the video segment has a time duration of less than four seconds, then a zero padding is applied on its spectrogram.

\begin{table*}[!t]
\vspace{-1mm}
\begin{center}
\caption{Performance comparison with EPIC Kitchens challenge baseline results. See text for more information.}
\label{tab:epicKitchensChallenge}
\resizebox{\textwidth}{!}{%
\begin{tabular}{ll|ccc|ccc|ccc|ccc}
                        &        & \multicolumn{3}{c|}{\textbf{Top-1 Accuracy}} & \multicolumn{3}{c|}{\textbf{Top-5 Accuracy}} & \multicolumn{3}{c|}{\textbf{Avg Class Precision}} &\multicolumn{3}{c}{\textbf{Avg Class Recall}} \\\cline{3-14}
                        &        & VERB      & NOUN      & ACTION     & VERB      & NOUN      & ACTION     & VERB   & NOUN  & ACTION & VERB   & NOUN  & ACTION \\\hline
\multirow{5}{*}{\rotatebox{90}{\textbf{S1}}}
                        &Chance/Random &12.62 &1.73 &00.22 &43.39 &08.12 &03.68 &03.67 &01.15 &00.08 &03.67 &01.15 &00.05\\
                        & TSN (RGB)      & 45.68     & {\bf36.80}& 19.86                & {\bf85.56}& {\bf64.19}& {\bf41.89}           & {\bf61.64}& 34.32     & 09.96                     & {\bf23.81}& {\bf31.62}& 08.81 \\
                        & TSN (FLOW)     & 42.75     & 17.40     & 09.02                & 79.52     & 39.43     & 21.92                & 21.42     & 13.75     & 02.33                     & 15.58     & 09.51     & 02.06 \\
                        & TSN (FUSION)   & {\bf48.23}& 36.71     & {\bf20.54}           & 84.09     & 62.32     & 39.79                & 47.26     & {\bf35.42}& {\bf10.46}                & 22.33     & 30.53     & {\bf08.83} \\
                        & Ours  & 34.26 & 08.60 & 03.28 & 75.53 & 25.54 & 11.49 & 12.04 & 02.36 & 00.55 & 11.54 & 04.56 & 00.89\\
                        \hline
\multirow{5}{*}{\rotatebox{90}{\textbf{S2}}}
                        & Chance/Random &10.71 &01.89 &00.22 &38.98 &09.31 &03.81 &03.56 &01.08 &00.08 &03.56 &01.08 &00.05\\
                        & TSN (RGB)      & 34.89     & 21.82     & 10.11                & {\bf74.56}& 45.34     &{\bf25.33}            & 19.48     & 14.67     & 04.77                     & 11.22     & 17.24     & 05.67 \\
                        & TSN (FLOW)     & {\bf40.08}& 14.51     & 06.73                & 73.40     & 33.77     & 18.64                & 19.98     & 09.48     & 02.08                     & {\bf13.81}& 08.58     & 02.27 \\
                        & TSN (FUSION)   & 39.40     & {\bf22.70}& {\bf10.89}           & 74.29     & {\bf45.72}& 25.26                & {\bf22.54}& {\bf15.33}& {\bf05.60}                & 13.06     & {\bf17.52}&{\bf05.81} \\
                        & Ours   & 32.09 & 08.13 & 02.77 & 68.90 & 22.43 & 10.58 & 11.83 & 02.92 & 00.86 & 10.25 & 04.93 & 01.90 \\

\end{tabular}}
\end{center}
\vspace{-5mm}
\end{table*}

We extracted the audio from all video segments using a sampling frequency of $16$ KHz\footnote{We used FFMPEG for the audio extraction \href{http://www.ffmpeg.org}{http://www.ffmpeg.org}.}, as it covers most of the band audible to the average person \cite{Heffner2007HearingRanges}. Since the audio from the videos have two channels, we joined them in one signal by computing their mean value. From this signal, we computed its short-time Fourier transform (STFT) \cite{McFee2015librosaAA}, as we are interested in noises rather than in human voices. The STFT used a Hamming window of length equal to roughly 30 ms with a time overlapping of $50\%$. For convenience of the input size of our CNN, we used a sampling frame length of 661. The spectrogram was equal to the squared magnitude of the STFT of the signal. Subsequently, we obtained the logarithm value of the spectrogram in order to reduce the range of values. Finally, all the spectrograms were normalized. The size of the input spectrogram image is $331\times248$.

\section{Experiments}

The objective of our experiments was to determine the classification performance by leveraging the audio modality on an egocentric action recognition task. We used the EPIC Kitchens dataset \cite{Damen2018EPICKITCHENS} on our experiments. Each video segment in the dataset shows a participant doing one specific cooking related \textit{action}. A labeled action consists of a \textit{verb} plus a \textit{noun}, for example, ``cut potato'' or ``wash cup''. We used the accuracy as a performance metric in our experiments. Moreover, as a comparison with using only visual information, we show the results obtained in the official test split of the dataset.

\paragraph{Dataset}
The EPIC Kitchens dataset includes 432 videos egocentrically recorded by 32 participants in their own kitchens while cooking/preparing something. Each video was divided into segments in which the person is doing one specific \textit{action} (a \textit{verb} plus a \textit{noun}). The total number of verbs and nouns categories in dataset is 125 and 352, correspondingly. Currently, this dataset is been used on an egocentric action recognition challenge \footnote{\href{https://epic-kitchens.github.io/2018}{https://epic-kitchens.github.io/2018}}. Thus, we only used the labeled training split and filtered it accordingly to \cite{Damen2018EPICKITCHENS}, i.e. only using the verbs and nouns that have more than 100 instances in the split. This results in 271 videos having 22,018 segments, and 26 verb and 71 noun categories. The resulting distribution of action classes is highly unbalanced.

\begin{figure}[!t]
\begin{center}
\includegraphics[width=\columnwidth]{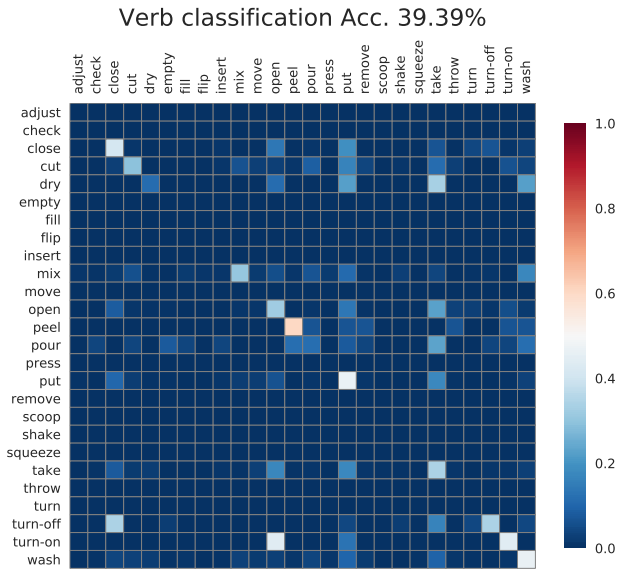}
\caption[]{Normalized verb confusion matrix.}
\label{fig:verbConfusionMatrix}
\end{center}
\end{figure}

\begin{table}[]
\centering
\resizebox{0.8\columnwidth}{!}{%
\begin{tabular}{l|c|c|c|}
\cline{2-4}
       & \multicolumn{1}{|p{0.2\columnwidth}|}{\centering Verb} & \multicolumn{1}{|p{0.2\columnwidth}|}{\centering Noun} & \multicolumn{1}{|p{0.2\columnwidth}|}{\centering Action} \\ \hline
\multicolumn{1}{|l|}{Chance/Random} & 13.11\% & 2.48\% & 0.75\% \\ \hline
\multicolumn{1}{|l|}{Ours (Top-1)} & 39.39\% & 13.41\% & 10.16\%  \\ \hline
\multicolumn{1}{|l|}{Ours (Top-5)} & 81.99\% & 35.60\% & 26.36\%  \\ \hline
\end{tabular}
}
\caption{Action recognition accuracy for all our experiments.}
\label{tab:actionRecognition}
\end{table}

\begin{figure*}[!t]
\begin{center}
\includegraphics[width=1.5\columnwidth]{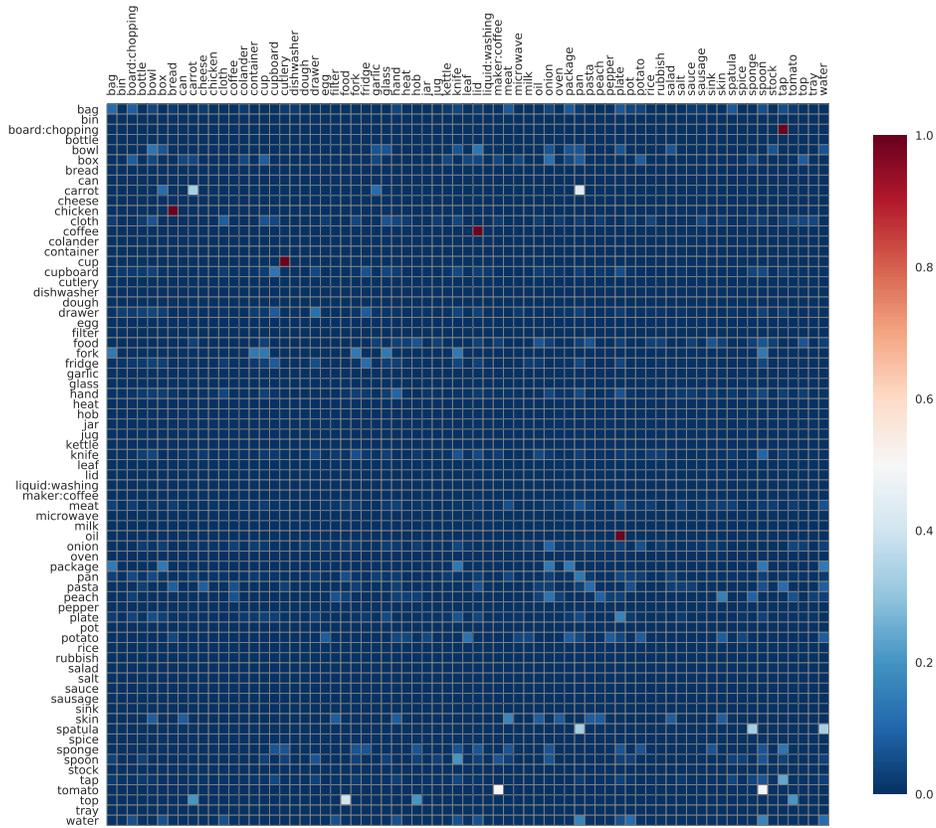}
\caption[]{Normalized noun confusion matrix.}
\label{fig:nounConfusionMatrix}
\end{center}
\end{figure*}

\paragraph{Dataset split}
For the different combinations of experiments (\textit{verb}, \textit{noun}, and \textit{action}), we divided the given labeled data into training, validation, and test splits considering all the participants. In all of the combinations, the data proportions for the validation and test splits were 10\% and 15\%, respectively. For the \textit{verb} and \textit{noun} experiments, the splits were obtained by randomly stratifying the data because each category has more than 100 samples. In the case of the \textit{action} experiment, the imbalance of the classes were considered for the data split as follows. At least one sample of each category was put in the training split. If the category had at least two samples, one of them went to the test split. The rest of the categories were randomly stratified on all splits.

\paragraph{Training}
For all our experiments we used the stochastic gradient descent (SGD) optimization algorithm to train our network. We used a momentum and a batch size equal to 0.9 and 6, correspondingly. The specific learning rates and number of training epochs for each experiment were: for \textit{verb} were $5\times10^{-6}$ during 79 epochs; for \textit{noun} were $2.5\times10^{-6}$ during 129 epochs; and for \textit{action} were $1.75\times10^{-6}$ during 5 epochs.

\paragraph{Results and discussion}

The accuracy performance for all experiments is shown in Table \ref{tab:actionRecognition}. Additionally, it also presents random classification accuracy baseline from the dataset splits described above. We calculated the random classification accuracy of $N$ categories as

\begin{equation}
acc=\sum_{i}^{N}{p^{train}_{i}\cdot p^{test}_{i}}
\end{equation}

\noindent where $p^{train}_{i}$ and $p^{test}_{i}$ are the occurrence probability for class $i$ in the train and test splits, accordingly. As means for comparison, we also present the tests results on the seen (S1) and unseen (S2) splits from the EPIC Kitchens challenge in Table \ref{tab:epicKitchensChallenge}. This results were obtained using the trained networks for the experiments on \textit{verb} and \textit{noun}.

The overall results indicate a good performance using audio alone for \textit{verb} classification. Additionally, the models fail to recognize categories that do not produce sound such as \textit{flip} (verb) and \textit{heat} (noun), as seen on the confusion matrices in Fig. \ref{fig:verbConfusionMatrix} and \ref{fig:nounConfusionMatrix}. In the case of verbs, the model also fails on conceptually closed verbs like the pairs \textit{turn-on}/\textit{open} and \textit{turn-off}/\textit{close}. In the case of \textit{noun} classification, the model incorrectly predicts objects that have similar materials, for example, the categories \textit{can} and \textit{pan} are metallic.

We consider that an object may have different sounds depending on how it is manipulated, and this may help to better discriminate the \textit{verb} performed on the object that may result visually ambiguous from an egocentric perspective. For instance, knifes and peelers are visually similar objects that could led to an action misclassification on verbs \textit{cut} and \textit{peel} used on nouns like \textit{carrot}, but their sounds were mostly correctly classified as seen in the confusion matrices in Fig. \ref{fig:verbConfusionMatrix} and \ref{fig:nounConfusionMatrix}. 

\section{Conclusion}

We presented an audio based model for egocentric action classification trained on the EPIC Kitchens dataset. We analyzed the results on the splits we made from the training set and made a comparison with the visual test baseline. The obtained results show that audio alone achieves a good performance on \textit{verb} classification (34.26\% accuracy). This suggests that audio could complement visual sources on the same task in a multimodal manner. Further work will directly focus on this research line.

{\small
\bibliographystyle{ieee_fullname}

}

\end{document}